\documentclass[journal]{IEEEtran}
\ifCLASSINFOpdf
   \usepackage[pdftex]{graphicx}
   \DeclareGraphicsExtensions{.pdf,.jpeg,.png}
\else
\fi
\usepackage{amsmath}
\usepackage{algorithmic}
\usepackage{algorithm}
\usepackage{array}
\usepackage{fixltx2e}
\usepackage{stfloats}
\usepackage{url}
\usepackage{csquotes}
\begin{document}
\title{Cross modal video representations for weakly supervised active speaker localization}

\author{Rahul~Sharma, Krishna~Somandepalli,
        and~Shrikanth Narayanan,~\IEEEmembership{Fellow,~IEEE}
        \thanks{Rahul Sharma (raul.sharma@usc.edu) and Shrikanth Narayanan (shri@ee.usc.edu) is with Electrical and Computer Engineering, University of Southern California, USA}
        \thanks{Krishna Somandepalli is with Google Inc.}}%<-this % stops a space
% \thanks{Rahul Sharma and Shrikanth Narayanan is with Department of Electrical and Computer Engineering, University of SOuthern California, USA}%
% \thanks{M. Shell was with the Department
% of Electrical and Computer Engineering, Georgia Institute of Technology, Atlanta,
% GA, 30332 USA e-mail: (see http://www.michaelshell.org/contact.html).}% <-this % stops a space
% \thanks{J. Doe and J. Doe are with Anonymous University.}% <-this % stops a space
% \thanks{Manuscript received April 19, 20/05; revised August 26, 2015.}}


\markboth{IEEE transactions on Multimedia}%
{Shell \MakeLowercase{\textit{Sharma et al.}}: Cross modal active speaker localization}



\maketitle

\begin{abstract}
An objective understanding of media depictions, such as inclusive portrayals of how much someone is heard and seen on screen such as in film and television,  requires the machines to discern automatically who, when, how, and where someone is talking, and not. Speaker activity can be automatically discerned from the rich multimodal information present in the media content. This is however a challenging problem due to the vast variety and contextual variability in the media content, and the lack of labeled data. In this work, we present a cross-modal neural network for learning visual representations, which have implicit information pertaining to the spatial location of a speaker in the visual frames. Avoiding the need for manual annotations for active speakers in visual frames, acquiring of which is very expensive, we present a weakly supervised system for the task of localizing active speakers in movie content. We use the learned cross-modal visual representations, and provide weak supervision from movie subtitles acting as a proxy for voice activity, thus requiring no manual annotations. We evaluate the performance of the proposed system on the AVA active speaker dataset and demonstrate the effectiveness of the cross-modal embeddings for localizing active speakers in comparison to fully supervised systems. We also demonstrate state-of-the-art performance for the task of voice activity detection in an audio-visual framework, especially when speech is accompanied by noise and music. 
\end{abstract}


% \fi
\IEEEpeerreviewmaketitle

\section{Introduction}
\label{sec:intro}
\IEEEPARstart{T}{remendous} variety and amounts of multimedia content are created, shared, and consumed everyday, and across the world, with a great influence on our everyday lives. These span various domains, from entertainment and education to commerce and politics, and in various forms; for example, in the entertainment realm these include film, television, streaming, and online media forms. 
There is an imminent need for creating human-centered media analytics to illuminate the stories being told by using these various content forms to understand their human impact: both societal and economic.
Recent efforts to address this need has led to the emergence of \textit{computational media intelligence} (CMI) \cite{cmi} which deals with building a holistic understanding of persons, places, and topics involved in telling stories in multimedia, and how they impact the experiences and behavior of individuals and society at large.

Creating such rich media intelligence requires the ability to automatically process and interpret large amounts of media content across modalities (audio, video, language, etc.), each modality with its strengths and limitations to help understand the story being told. %These modalities inherently have vast variety, heterogeneity, dynamic variability, and complex relations between them, and offer only partial information about the \textit{what-why-how-where-when} type constructs of interest. 
The ability to process multiple modalities hence becomes essential to learn robust models for media content analysis. It should be noted that humans concurrently process and experience different aspects of the presented media: sights, sounds, and language use to develop a holistic understanding of the story presented \cite{klemen2012current}.  For example, several studies in psychology and neuroscience have shown evidence for how visual perception in humans is intertwined with other senses such as sound and touch. These mechanisms can be altered even at early stages of development of the primary visual cortex (e.g.,~\cite{shams2010crossmodal}). 
This integration of multiple sensory modalities to holistically perceive visual stimuli is a widely studied field in human psychology, referred to as \textit{crossmodal perception}~\cite{schmiedchen2012crossmodal}. Recently, there have been several works focused on the computationally harnessing the idea of crossmodal perception in the audio-visual domain. Most of  these studies use the idea of the naturally existing relations in the audio and the corresponding visual frames, in produced media content~\cite{arandjelovic2017look,objectsthatsound,multisensory_ownes,sound_pixel}.

\emph{When} and \emph{where} constructs are the fundamental pillars of CMI, for developing a holistic understanding of a scene, which direct to locate the action of interest in time and space. In this paper, we address the problem of audio-visual speech event localization in (Hollywood) movies, which essentially detects the audio speech event in time (\emph{when}) and the corresponding speech activity in the visual frames (\emph{where}). Inspired by the cross-modal integration in humans to address the challenges of partial observability and dynamic variability of the audio and visual modalities, we developed a cross-modal neural network that can efficiently fuse the complementary information of the visual and audio modalities to effectively localize an audio-visual speech event.

% We trained a deep neural network to learn visual representations with a cross-modal objective, where the system is designed to predict audio voice activity detection (VAD) operating on just the visual information. We employed a multiscale temporal convolutional network consisting of 3D convolutional neural networks (CNNs) and stacked convolutional LSTM networks. Furthermore we setup a multiple instance learning (MIL) task for active speaker localization (in visual frames) in a weakly supervised fashion, utilizing the learned cross-modal representations and availing the weak supervision from audio VAD labels. For the target use case of movie content analysis, we obtain the required coarse speech activity labels from the movie subtitles, employing no manual annotations.

% The contributions of this work include: i) a cross-modal problem formulation for learning visual representations for visual speech event localization using a proxy objective of audio speech event detection; ii) a modified version of hierarchichal context aware (HiCA \cite{icip}) architecture  with stacked convolutional Bi-LSTMs to provide multi-scale temporal context to learn the required visual representations; iii) an end-to-end trainable weakly supervised setup for the task of active speaker localization; and iv) a state-of-the-art performance for the task of audio speech event detection when speech is accompanied with noise or music. 
In our preliminary work \cite{icip}, we introduced a cross-modal problem formulation for the task of visual voice activity detection. We proposed a 3D convolutional network that observes the raw visual frames of a video segment and predicts the posterior for segment-level audio voice activity detection (VAD). We further established that the learned embeddings were capable of localizing humans in the visual frames. In this work we further advance the proposed framework for audio-visual speech event localization, where audio speech event refers to the voice activity detection in time (audio modality) and visual speech localization refers to localizing active speakers in space (visual frames). The novel contributions reported include the following:

\begin{enumerate}
    \item We introduce an enhanced cross-modal architecture consisting of 3D convolutional neural networks (CNNs) and stacked convolutional Bi-LSTMs. This enables the system to capture multi-scale temporal context and introduces an ability to learn hierarchical abstractions in the presented information. The presence of convolutional operations throughout the architecture, in CNNs as well as in Bi-LSTMs, enables the system to preserve the spatiotemporal information, thus making it interpretable at several levels.
    \item We present an end-to-end trainable cross-modal system for active speaker localization in visual frames, trained in a weakly supervised fashion. The proposed setup utilizes a multiple instance learning formulation designed for detecting the presence of speech in audio while considering the location of active speaker faces in the visual modality as the key instances. Furthermore, we evaluate the system on a public dataset and demonstrate performance comparable to state-of-the-art fully supervised methods. 
    \item We present an audio-visual system for segment-level voice activity detection (VAD) which utilizes a fusion of the learned cross-modal visual representations and state-of-the-art audio representations. We conduct experiments on public benchmark datasets to show the state-of-the-art VAD performance for the cases when speech is accompanied by noise or other interference.
\end{enumerate}

% The rest of the paper is organized as follows. In Section~\ref{sec:related work} we discuss the background research and related works on cross-modal learning and active speaker localization. In Section~\ref{sec:methods} we describe the methodological aspects of the cross-modal problem formulation and the implementation details for the end-to-end setup. We present the evaluation strategy and the experimental results in Section~\ref{sec:expt}. We conclude by summarizing and listing future directions in Section~\ref{sec:conclusion}.
\section{Related Work}
\label{sec:related work}
\subsection{Cross-modal learning}
% There has been a recent surge of studies focused on cross-modal machine perception, especially in  media content analysis. The idea of cross-modal learning primarily revolves around modelling one modality guided by another. In \cite{somandepalli2018multimodal}, the authors target video advertisement classification, using cross-modal learning, where audio-visual representations are modeled using cross-modal autoencoders, structured to reconstruct one modality from the other. In a more recent work by \cite{cross-modal(CMRAN)}, a cross-modal relation-aware network is proposed for audio-visual event localization. The self-attention mechanism is modified such that the query is derived from one modality while the key-value pairs are derived from the other. Working along similar lines, \cite{cross-modal(fake_news)} targets the problem of fake news detection using a cross-modal residual network, where the text modality guides the attention for learning visual representation and vice-versa.

There has been a recent surge of studies focused on cross-modal machine perception, especially in  media content analysis. The idea of cross-modal learning primarily revolves around modelling one modality guided by another. In~\cite{somandepalli2018multimodal}, the authors target video advertisement classification, using cross-modal autoencoders, reconstructing one modality from the other. In a more recent work by~\cite{cross-modal(CMRAN)}, a cross-modal relation-aware network is proposed for audio-visual event localization involving a self-attention mechanism where query is derived from one modality while the key-value pairs the other. Another work~\cite{cross-modal(fake_news)} targets the problem of fake news detection using a cross-modal residual network, where the text modality guides the attention for learning visual representation and vice-versa. In our earlier work~\cite{icip}, we proposed a cross-modal problem setup for the the task of visual VAD invovling a hierarchically context-aware network (HiCA) which observes the visual frames and predict the audio VAD labels. 

% In our earlier work \cite{icip}, a hierarchically context-aware network (HiCA) was proposed for an audio-centric VAD, but derived from the visual modality.  The learned cross-modal embeddings were systematically analyzed and results showed that such a network can localize human faces. The present work is primarily inspired by the HiCA network. We advance this formulation by modifying the architecture and formalizing a weakly supervised setup for active speaker localization in the visual modality. 

\subsection{Weakly supervised object detection (WSOD)}
WSOD refers to the training setup when only image level labels are provided for supervision opposed to bounding box labels in fully-supervised scenarios. Recent research in WSOD can be broadly categorized into two directions, $i)$ Class activation maps (CAMs), and $ii)$ Multiple instance learning (MIL) based setups. CAMs based methods leverage the relationship between CNN embeddings and the class posteriors to compute localization maps. One of the earlier approaches~\cite{WSOD_selfthaught} used the idea that the recognition score will drop if the object of interest is artificially masked out in the input image. The idea of CAMs~\cite{WSOD_cams} was initially proposed to compute the discriminative image regions for a class of interest in the case of linear prediction layers. Grad CAM~\cite{Grad-CAM} was later introduced, generalizing CAMs by using the gradients of the posteriors with respect to the activations of the pertinent layer. Furthermore, GradCAM++~\cite{WSOD_gradcam++}, introduced weighted average of pixel-wise gradients to improve the coverage of detections and dealt with multiple occurrences of the same object.

MIL setups pose the input image for classification as a bag of instances where instances are object proposals. In an early attempt~\cite{WSOD_wsddn} a two stream CNN was proposed, one stream to predict bag scores while the other one to compute the instance level scores. Recently~\cite{WSOD_midn} proposed a multistage instance classifier (MIDN) to predict the tighter object detection boxes, which is further enhanced to improve coverage of detection by using 2 MIDMs~\cite{WSOD_cmidn}. To alleviate the non-convexity issues associated with MIL~\cite{WSOD_cmil} proposed to use a combination of smoothed loss functions. 
% Furthermore, \cite{WSOD_activity} proposed to learn the spatial priors of objects with respect to humans in different actions to reduce the search space for the objects.

\subsection{Active speaker localization}
Earlier works~\cite{everingham2006hello} in active speaker detection largely focus on using the activity in the lip region available in the visual modality. In another approach~\cite{outoftime}, authors proposed to use the synchrony between the cropped images of lip regions and the associated audio to determine active speakers. Furthermore, ~\cite{JayChakroborty} introduced the use of cues from upper body motion to determine an active speaker, which they further refined using personalized voice models~\cite{jayactive}. Recently~\cite{avaActiveSpeaker} proposed a large scale dataset (AVA active speaker dataset), consisting of movies and the corresponding active speaker annotations along with baseline performance using a supervised framework. Several frameworks have since followed~\cite{ava_multitask, ava_naver, ava_asdcontext} for improving the performance on the AVA dataset. But all these works are restricted to supervised frameworks. To overcome the need of expensive annotations~\cite{asd_selfsupervised} proposed a self-supervised framework trained for the task of audio visual correspondence using the optical flow information.
% to construct audio-visual objects. 
% In this work, we propose a weakly supervised framework acquiring the weak supervision from VAD labels and compare the performance against the supervised frameworks on the AVA dataset.
\subsection{Sound source localization}
The problem of active speaker localization falls within the general domain of sound source localization, but for a particular audio event: speech. The core idea driving the research in this direction is to exploit the existing audiovisual correspondence in the media content. Earlier efforts~\cite{ssl_old1,ssl_harmony, ssl_old2} used canonical correlation analysis to model the audio-visual correspondence. Recent research has been dominated by self-supervised deep learning methods, where researchers try to capture the audio-visual correspondence using various proxy tasks. One such proxy task~\cite{sound_pixel, ssl_soundmotion} uses the additive nature of audio and reconstruct the sound for each pixel by learning a mask for the audio spectrogram. Another proxy task~\cite{multisensory_ownes} predicts the time alignment of the given audio and video pair. The work by~\cite{objectsthatsound} used the audio-visual correspondence to predict a localization score for every pixel and~\cite{ssl_od} extended the same formulation for object detection. Furthermore~\cite{cross-modal(CMRAN)} proposed a cross-modal attention mechanism for audio event classification and used the learned attention for modeling the localization task. Majority of these works qualitatively established the gained localization ability from the inherent audio-visual correspondence but lacks quantitative evaluation. In this work we present a qualitative as well as thorough quantitative analysis of the acquired localization ability of the visual embeddings. 
\section{Methodology}
\label{sec:methods}
In this section, we first introduce the problem formulation for learning cross-modal visual representations, followed by the cross-modal neural network architecture and relevant implementation details. We further elaborate on the methods developed to formally use the learned visual representations for active speaker localization in a weakly supervised manner. The overview of the complete end to end framework is given in Fig~\ref{fig:MIL_setup}.

\subsection{Problem Description}
\label{subsec: problem_formulation}
\begin{figure*}[tb]
    \centering
    \includegraphics[width=0.85\textwidth,keepaspectratio]{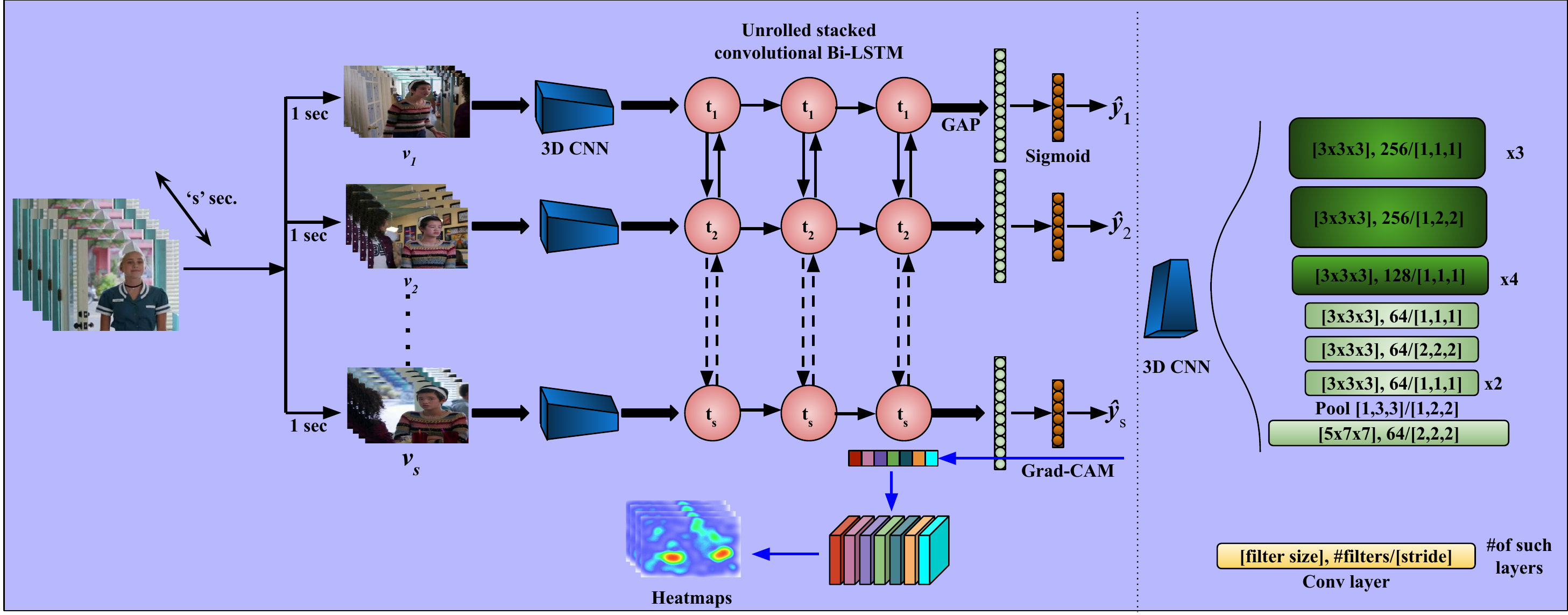}
    \caption{The crossmodal architecture with 3D CNNs and stacked convolutional BiLSTM layers}
    \label{fig:crossmodal_arch}
\end{figure*}
The work in this paper is especially motivated by the application of active speaker localization in media content such as entertainment media, notably Hollywood movies. From a computer vision perspective, movie videos are challenging due to the presence of rich variety and high dynamics in the content with potentially multiple variable number of persons in both the foreground and the background.  Supervised modeling of such videos requires large amounts of (labeled) data. In particular, training an audiovisual system for person localization task in a supervised fashion requires large-scale bounding box annotations, which are tedious and  expensive to acquire. Inspired by the recent success of cross-modal representations in understanding media content~\cite{cross-modal(CMRAN), cross-modal(fake_news)}, we formulate our problem in a cross-modal fashion where we model a function of audio modality i.e., talking/non-talking person, by directly observing the visual frames. This helps us in circumventing the widespread issues of drop in performance while jointly modeling multiple modalities against uni-modal systems~\cite{training_hard}, that arises primarily due to the difference in the rate of generalization for different modalities.

In our preliminary work~\cite{icip}, we trained a cross-modal network for predicting segment-level audio voice activity by using the  visual information and established that the learned embeddings implicitly acquired a capability to localize humans in the visual frames. Motivated by the attained localizing ability, in this work we modified the cross-modal formulation described in \cite{icip}, such that the learned embeddings can localize active speakers in the visual frames. To do so, we propose a modified formulation of the learning task to predict the \emph{presence of speech} (PoS) for a video segment by observing the visual frames. For a given video segment $v_i$ of $t-seconds$, we define PoS as the step function of the duration of voice activity.
\begin{align}
    \text{PoS} =
        \begin{cases}
            1 & \text{duration of voice activity} > 0 \\
            0 & \text{duration of voice activity} =0
        \end{cases}
\end{align}
The task of predicting PoS is specifically chosen with a hypothesis that the neural network will assess the active speaker regions in visual frames as the most salient to detect the PoS in the video segment. In our experimental setup we use data from Hollywood movies for training under this formulation, and thus utilize the readily available movie subtitles to acquire the PoS labels involving no manual annotations. Since obtaining a segment-level (video segment of t-sec) PoS label is a more relaxed scenario against obtaining a segment-level voice activity label, it enables us to obtain finer labels as compared to the strategy introduced in \cite{icip}.

Formally, given a video $V$, we partition the video into smaller segments $v_i$ of $t-seconds$ each. For each of the $v_i$, we acquire a label $y_i$, where $y_i$ indicating the PoS in the video segment. The network sees $k$ such small segments at once, and the network is trained for the mapping problem $v_i \rightarrow y_i$. In the current setup, $t=1sec$ and $k=10$.
\begin{align}
    % \begin{split}
    % V = \{v_1, v_2, \dots v_K\}\quad
    % Y = \{y_1, y_2, \dots y_K\}\quad y_i \in \{0,1\} \\
    \{v_i,\dots, v_{i+k}\} \rightarrow \{y_i, \dots, {y_{i+k}}\} \quad y_i \in \{0,1\}
    % \end{split}
\end{align}
\subsection{Cross-modal network architecture}
\label{subsection: crossmodal_architecture}
\subsubsection{Architecture}
To model the visual signal in a cross-modal fashion, in preliminary work~\cite{icip}, we introduced a Hierarchical Context-Aware (HiCA) architecture providing the temporal context at different levels, modeling the short-term context using 3D CNNs and long-term context using BiLSTM. Furthermore, we qualitatively and quantitatively established that the trained representations were selective to human faces and the human body. In this work, we enhance the decentralized temporal context of the HiCA architecture by employing three stacked convolutional Bi-LSTM on top of the 3D CNNs to provide multi-scale temporal context. The introduction of the stacked Bi-LSTMs is  motivated by the fact that stacked LSTM networks introduce a hierarchical level of abstractions, as established in various works~\cite{hermans2013training} in the field of Natural Language Processing. The convolutional Bi-LSTMs enable the integrated interpretability for the architecture, since they preserve the spatial and temporal structure of the input. Such a model also enables visualizing the learned representations at different levels of the stacked Bi-LSTMs, allowing to analyze the learned hierarchical abstractions. The elaborated neural network architecture is shown in Figure~\ref{fig:crossmodal_arch}.

\subsubsection{Experimental implementation details}
\label{subsection:subtitles}
The network is trained on a set of 268 Hollywood movies, released during the period 2014-18. The videos are sampled at 24 frames per second and are lowered in resolution to 180 x 360 pixels. The \emph{Presence of speech} labels are implicitly obtained using the readily available movie subtitles since they correspond to the  human speech dialogues present in the movies. The obtained labels are coarse and do not employ any manual annotations. The subtitles are first processed to remove the presence of special sounds by removing the content quoted within $[.]/\{.\}$.  It has been observed that the acquired subtitles are not accurately time aligned with the audio. We used the gentle force aligner~\footnote{https://lowerquality.com/gentle/}, a Kaldi-based~\footnote{https://kaldi-asr.org/} tool to align speech and text, which time aligns the subtitles and audio and provides a confidence score with each alignment. We discard the part of the videos which has not been aligned with high enough confidence (empirically determined). We further compute a binary label for each $t-sec$ of the video segments using the presence of subtitles as a proxy for presence of speech. To provide a tolerance for subtitle alignment errors, we assign a video segment a positive PoS label only if it has speech content for more than $10\%$ of the duration. 

After pre-processing and time aligning the subtitles with audio, we obtained, on average, nearly $70\%$ of the movie duration with a high enough speech-subtitle alignment confidence score. We used $k=10sec$ and $t=1sec$, which were driven heuristically, ensuring that CNNs and LSTMs observe enough temporal context to learn. Our training set consists of nearly 360 hours of video data, which comprises 130k samples ($1.3$ million video-label pairs, since each sample consists of 10 pairs). The network has been optimized to minimize the cross-entropy loss  using an accelerated SGD optimizer for nearly $1$ million iterations for a batch size of $8$. 
\begin{figure}
    \centering
    \includegraphics[width=0.45\textwidth,keepaspectratio]{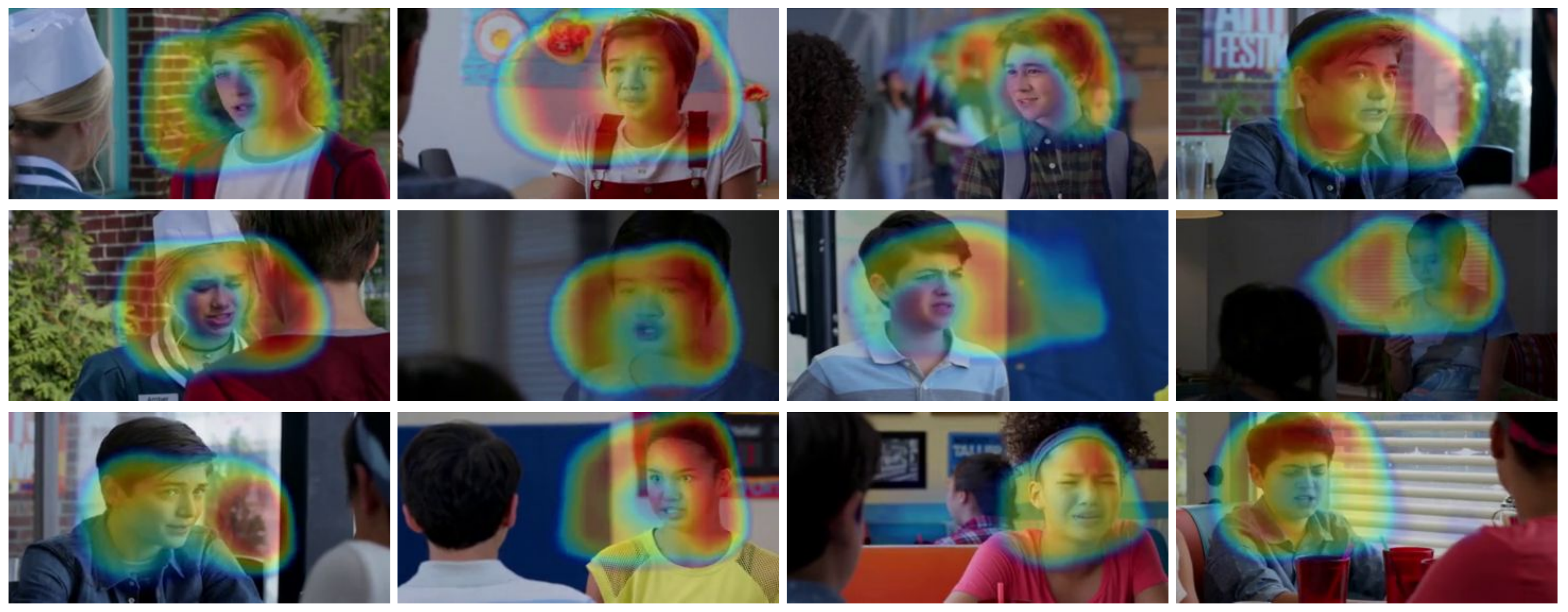}
    \caption{Class activation maps for positive class imposed on the input frames showing the localization ability of the learned embeddings.}
    \label{fig:single_person}
\end{figure}
\subsection{CAM-assisted active speaker localization}
\label{subsec:CAM_assistance}
The utmost factor motivating the use of convolutional networks throughout the cross-modal architecture is the ability of CNNs to enhance the interpretability of the learned embeddings. In this work we use an extension of GRAD CAMs~\cite{Grad-CAM} to 3D CNNs, as introduced in our preliminary work~\cite{icip} to visualize the information learned by the visual embeddings.  We first differentiate the output sigmoid score, the posterior $\hat{p}_i$ for PoS, with respect to each of the filters $F^m$ of the pertaining convolutional layer (the last conv-LSTM layer in current scenario) with $m$ filters. The obtained gradients are aggregated across temporal and spatial dimensions to obtain the contribution of each filter towards the presence of speech event. The filters of the convolutional layer in consideration are averaged in accordance with the weights computed in Eqn.~\ref{eq:cams}, and rectified linearly to obtain the final class activation maps, $C$.
\begin{align}
\label{eq:cams}
    \alpha_m &= \frac{1}{Z}\sum_i\sum_j\sum_k\frac{\partial {\hat{p}}}{\partial{F^{m}_{ijk}}} & C &= ReLU(\sum_m\alpha_mF^m)
\end{align}

We use this framework to quantitatively analyze the learned abstractions among the stacked LSTM layers.  Fig.~\ref{fig:single_person} shows the positive-class activation maps for selected key-frames from the video segments of the TV show \emph{Andi Mack},  not seen by the network earlier. Although the CAMs are designed for the purpose of enhancing the interpretation of the involved neural network, we observe in Fig.~\ref{fig:single_person} that the CAMs provide a non-trivial signal for localizing active speakers. We propose to extend the CAMs framework in a classification scenario, where for a given face bounding box, we use the CAMs derived signal to predict the status of the face as active speaker or not-speaking. Such a system will quantitatively validate the earlier presented hypothesis that the learned cross-modal representations for the task of PoS have implicit information for localizing active speakers. 

The proposed cross-modal system consists of 3DCNNS and ConvBiLSTMs, which preserve the temporal and spatial information throughout the network, and thus enable the CAMs to take advantage of the abstracted information in the later stacked BiLSTM layers. We compute the CAMs for the last ConvBiLSTM layer for a given video segment, part of a longer video clip. Since the ultimate layer in the network has a temporal resolution of 6 fps, and so are the obtained CAMs. We linearly interpolate the CAMs across the temporal and spatial dimensions to obtain the maps matching the spatio-temporal resolution of the given video segment. We normalize the obtained CAMs to restrict their value in range [0,1], by using a min-max normalization (using Eqn.~\ref{cam_norm} for a frame $f$) across the longer video clip. The min-max normalization is motivated by the assumption that there exists at least one instance of active speaker in the longer video clip, thus marking that as the benchmark for the CAM scores.
\begin{align}
    \label{cam_norm}
    \begin{split}
        C_f &= \frac{C_f}{Max(C) - Min(C)} \\
        s_j^f &= max C_f[y_1:y_2,x_1:x_2]
    \end{split}
\end{align}

\begin{figure*}[tb]
    \centering
    \includegraphics[width=0.85\textwidth,keepaspectratio]{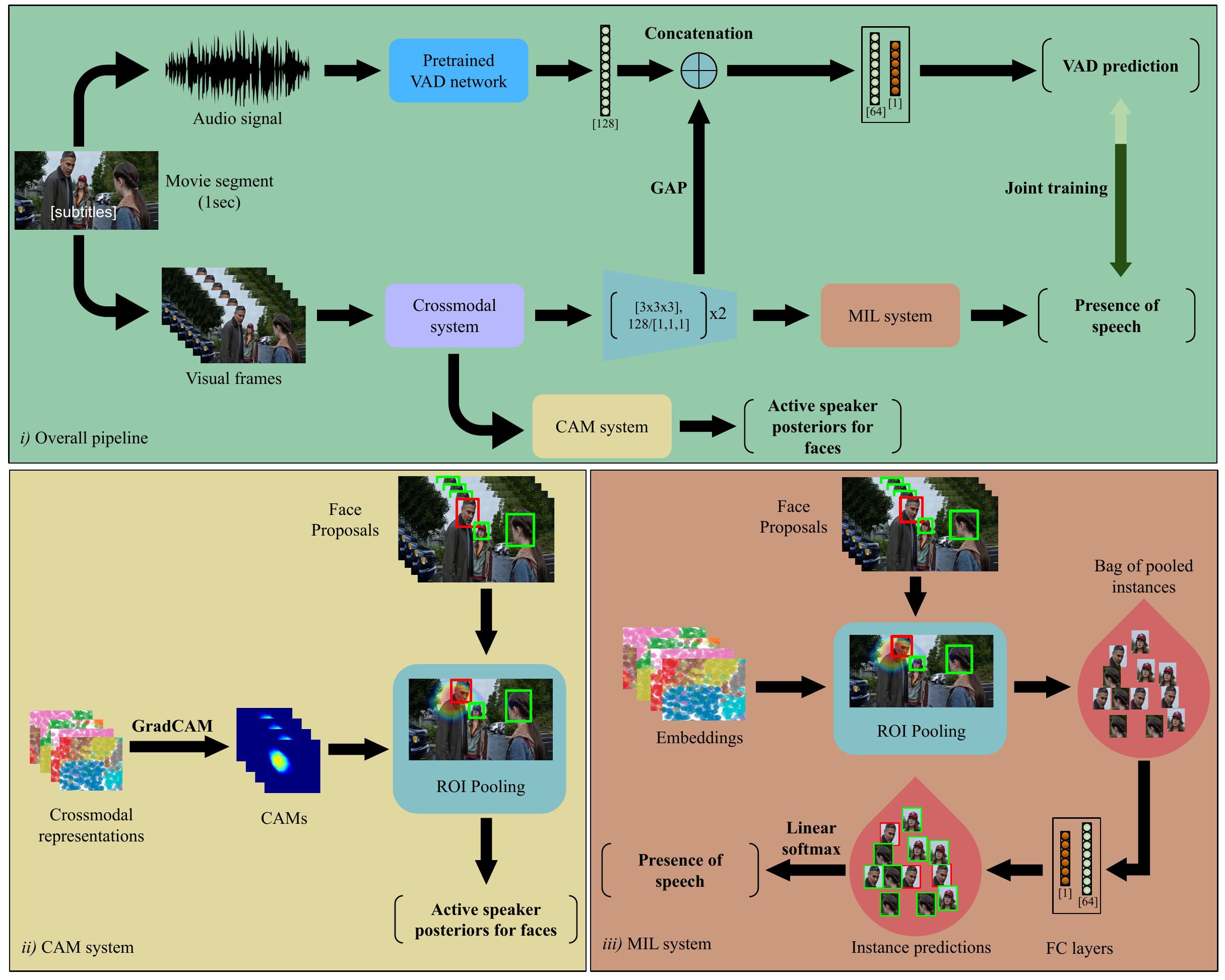}
    \caption{i) Overall training setup and building blocks for active speaker localization. ii) CAMs assisted setup processing the output of crossmodal CNN network to compute active speaker posteriors. iii) Weakly supervised MIL setup to predict Presence of Speech considering active speakers as key instances.}
    \label{fig:MIL_setup}
\end{figure*}

Formally, we denote a bounding box of the $j^{th}$ face in frame $f$, for a video segment $v_i$ as $b_j^f = [(x_1, y_2), (x_2, y_2)]$. We ROI pool the normalized CAM, $C_f$ for the frame $f$, pertaining to the face bounding box, and compute the posterior score, $s_j^f$ by max-pooling the pooled CAM. The framework is shown in the Fig.~\ref{fig:MIL_setup}-ii.
\subsection{Weakly supervised active speaker localization}
\label{subsec:wsod_network}
In the previous section, we used CAMS, a function of the embeddings, rather than the embeddings directly, to formalize a system for active speaker detection. Such a system helps us to explicitly validate our hypothesis, that the learned visual representations can localize active speakers in visual frames. In this section, we propose a systematic method to utilize the learned embeddings for active speaker localization, in a weakly supervised manner. We present a multiple instance learning (MIL) setup optimized for the proxy task of the presence of speech (PoS) by observing the learned cross-modal embeddings \S\ref{subsection: crossmodal_architecture}, and model the active speakers as the key instances. The setup is inspired by the recent works in weakly supervised object detection\cite{WSOD_cmidn, WSOD_cmil, WSOD_instance, WSOD_midn}. 

The problem of active speaker localization in space, is closely related to the active speaker detection in time, which is widely known as voice activity detection (VAD). The prevalent audio-based VAD systems are severely challenged by the variety and variability in the noise seen in real-world conditions. Various works have shown that visual features pertaining to localizing active speakers can complement the audio representations in noisy scenarios. We formalize a multimodal system for VAD using the learned cross-modal visual representations along with state-of-the-art~\cite{rajat_vad} audio representations and show the usefulness of the visual embeddings especially in the cases when speech is accompanied by noise.

\subsubsection{MIL problem formulation}
Multiple instance learning falls under the domain of supervised learning scenarios, where given labeled bags, each bag consisting of multiple instances, we learn a mapping from bags to labels. Particularly for a binary classification task, the bag is assigned a positive label if at least one of the instances in the bag is positive. This scenario fits appropriately with our problem formulation described in ~\S\ref{subsec: problem_formulation}. We define a small video segment, $v_t$, as a bag, and all the faces appearing during the video segment as instances. We train the system for the presence of speech (PoS) events, thus assigning $v_t$ (\emph{bag}) a positive label only if at least one of the faces (\emph{instances}) corresponds to the active speaker. 

The proxy task of PoS in the MIL setup is specifically chosen, so as to make it consistent with the earlier cross-modal setup (\S\ref{subsection: crossmodal_architecture}). With such a setup the cross-modal architecture observes the raw visual frames and is trained for the PoS labels. The output embeddings along with face proposals further become input for the MIL system, which is also trained for the PoS task. This consistency in both the setups' learning tasks makes them compatible to be trained in an end-to-end fashion. The combined CNN + MIL architecture observes the raw visual frames and predicts the PoS tags. Due to computational constraints, in this work, we restrict to training the two components separately. 
\subsubsection{Implementation details}
The MIL system observes the visual representations, obtained using the proposed cross-modal architecture (\S\ref{subsection: crossmodal_architecture}), and face detection boxes\footnote{Neven Vision \textit{f}R\textsuperscript{TM} API} for each frame to predict the PoS in the video segment. The face detections are sampled to match the temporal resolution of the visual representations. We employ an ROI~\cite{fastRCNN} pooling layer to generate instance-level descriptors, which next passes through a set of fully connected layers, terminating with a sigmoid activation layer. Thus, we obtain the instance level predictions, which are further pooled using a modified linear softmax~\cite{linear_softmax} to produce the bag level predictions. Since we froze the weights of the cross-modal (HiCA) architecture while training the MIL system, we introduce a trainable 3D-CNN block which observes the visual embeddings from the HICA architecture and, its output representations further act as an input to the MIL system. This enables fine tuning of the initially learned embeddings for the MIL system. The complete architecture is shown in Fig.~\ref{fig:MIL_setup}.

In a recent work~\cite{linear_softmax} it was suggested that in MIL system training, the max-pooling of the instance posteriors, to obtain the bag posteriors, shows a selective behavior highlighting one of the instances among all others. It was also pointed that linear softmax pooling boosts the larger posteriors while suppressing the smaller posteriors at the same time. For the application of active speaker localization, we assume the case of non-overlapping speakers, i.e., there can be at most one active speaker in each frame. This requires the selective behavior of the pooling method, selecting one instance (face), at the frame level. Concurrently, there will likely be more than one frame in the video, consisting of instances of active speakers. Thus we require linear-softmax kind of behavior at inter-frame level pooling, boosting the posterior of the more confident frames.

We propose to use a combination of the two pooling methods, thus pooling the instances in each frame using max-pooling to obtain the frame-level posteriors. We further pool the frame-level posteriors using linear softmax pooling operation to obtain the video level posterior score. Let the instance posterior for the $i^{th}$ face in frame $f$ is denoted as $\hat{P}_{fi}$. The bag posterior, $\hat{P}$, is obtained as shown in eq~\ref{eq:linear_softmax}. We optimize the MIL system for the cross-entropy loss, $Loss_{\text{MIL}}$ between the bag posteriors $\hat{P}$ and corresponding PoS labels.
\begin{align}
    \label{eq:linear_softmax}
    \hat{P} = \frac{\sum_f(\max_{i}\hat{P}_{fi})^2}{\sum_f(\max_{i}\hat{P}_{fi})}
\end{align}

Simultaneously, we train a late-fusion multi-modal network for the task of VAD, for the same set of input data. We obtain the VAD labels for the video segments using the movie subtitles in a similar fashion as we obtained the PoS labels, described in \S\ref{subsection: crossmodal_architecture}. We provide a positive binary label if the speech content is present more than 50\% of the duration of the video segment, and a negative label otherwise. We use the pre-trained state-of-the-art~\cite{rajat_vad} audio representations for the task of VAD, and pass them through an FC layer to further concatenate with the video representations, obtained from the introduced 3D CNN block between the HICA architecture and the MIL system. The concatenated representations are further passed through a set of FC layers to give a segment-level posterior score for VAD. We compute a cross-entropy loss, $Loss_{\text{av}}$ and optimize jointly with the  MIL system, minimizing the convex combination of the two losses (using $\alpha=0.9$). 
\begin{align}
    \label{eq: joint_loss}
    Loss_{\text{joint}} = \alpha*Loss_{\text{MIL}} + (1-\alpha)*Loss_{\text{av}}
\end{align}
\subsubsection{Post processing}
\label{subsec:post_processing}
We propose post-processing steps to refine the obtained instance-level posteriors by imposing audio and visual constraints.
\begin{itemize}
    \item \emph{Video post processing}: We assume that at any point in time there exists at most one speaker. To apply such a constraint, we penalize the posteriors of all the faces, but the one with the maximum posterior score, in every frame. This helps in removing the false positives in the predictions. 
    \begin{align}
       \hat{P}_{fi} =
        \begin{cases}
            \hat{P}_{fi}, & \hat{P}_{fi} = \max_{i}\hat{P}_{fi}. \\
            \frac{\hat{P}_{fi}}{\beta}, & \text{otherwise}.
        \end{cases}
    \end{align}
    
    \item \emph{Audio post processing}: We use the fact that if there is no speech detected in the audio modality, there can not be any speaker present in the visual frames. To integrate this information in the active speaker posteriors, we impose a penalty on all the faces in the frames appearing in non-voice activity regions. Essentially, 
    \begin{align}
        \hat{P}_{fi} = 
        \begin{cases}
            \hat{P}_{fi}, & \text{if frame $f$ in VAD region}.\\
            \frac{\hat{P}_{fi}}{\gamma}, & \text{otherwise}.
        \end{cases}
    \end{align}
 \end{itemize}
\section{Experiments and evaluations}
\label{sec:expt}
% In this section, we first present the qualitative evaluation of the learned cross-modal visual representations. We further present a quantitative evaluation of the proposed systems for active speaker localization, the CAMs-assisted, and the MIL-based weakly supervised system. We provide details about the evaluation datasets, the experimental setups, and the metrics reported. Furthermore, we compare the performance of the proposed system against baselines and state-of-the-art systems.
In this section, we present the qualitative and quantitative evaluation of the learned cross-modal visual representations by evaluating the proposed systems for active speaker localization, the CAMs-assisted, and the MIL-based weakly supervised system. We provide details about the evaluation datasets, the experimental setups, and the metrics reported. 

\subsection{Qualitative analysis}
\label{subsec: qualitative}
Here we evaluate the hypothesis, that the visual embedding learned to detect the presence of speech in audio modality can localize active speakers in the visual frames, from a qualitative perspective. We visualize the salient regions in the video frames for the positive class i.e., presence of speech event in the audio modality using the methodology described in ~\S\ref{subsec:CAM_assistance}. Fig.~\ref{fig:single_person} shows the CAMs imposed on the frames in form of heatmaps and demonstrates that the positive class activations situate around human faces with high concentration. 

\begin{figure}
    \centering
    \includegraphics[width=0.45\textwidth,keepaspectratio]{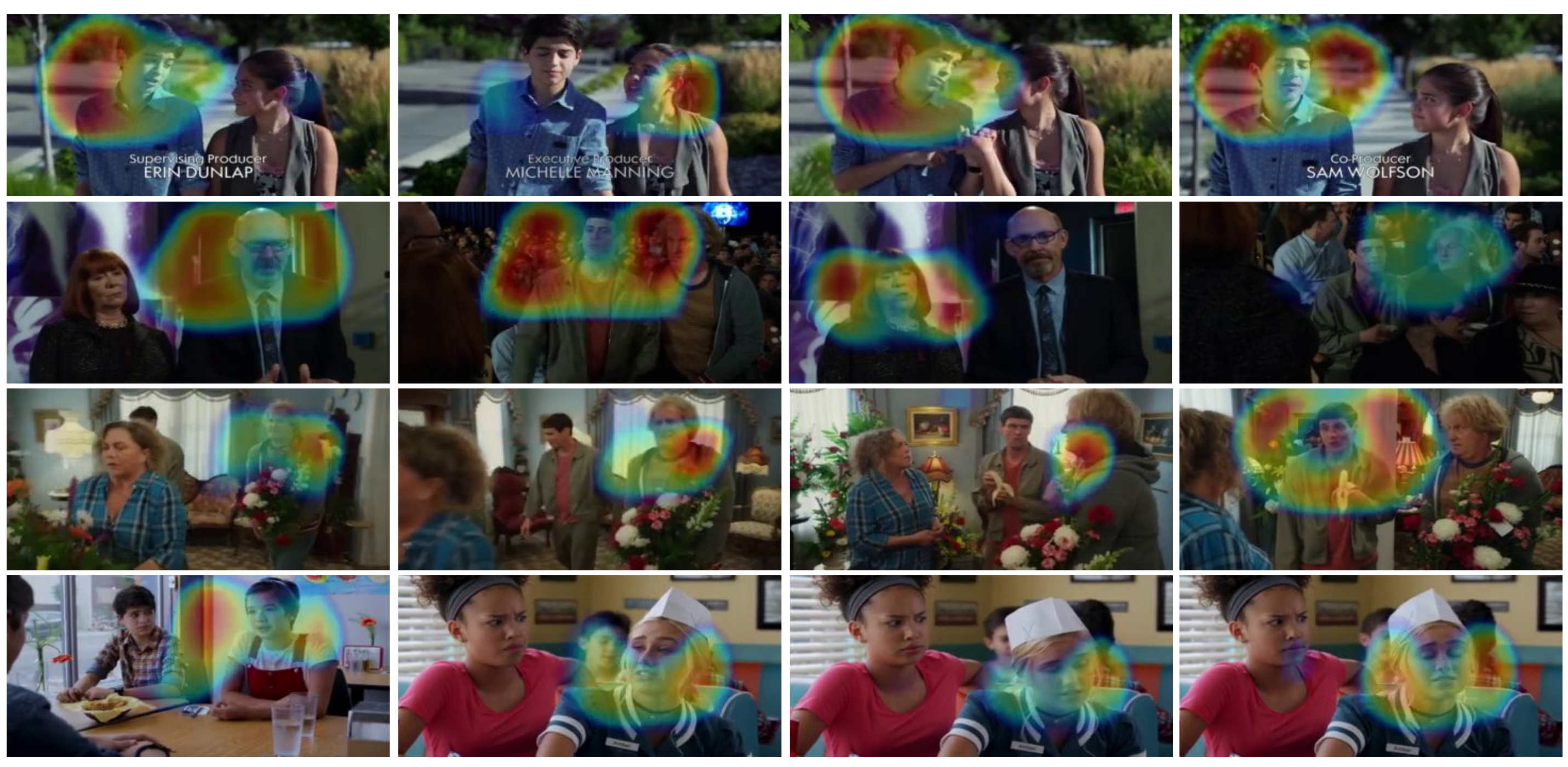}
    \caption{Illustration of localization performance of the crossmodal embeddings specifically for the frames with more than one face.}
\label{fig:multi_person}
\end{figure}

\textbf{\textit{Multiple faces in a frame:}} An approach for generalized sound source localization was recently proposed by \cite{multisensory_ownes}, where a neural network for audio-visual synchrony was trained, a framework that appears to be closest to our case. This work presents class activation maps for various videos in audio-set~\cite{audio-set} and shows that the learned audio-visual representations are selective to human faces and moving lips in case speech event. But the majority of cases presented in this work consist of just one human face in the frame; moreover, they do not provide extensive quantitative analysis to support the claim. As we specialize the cross-modal embeddings for the presence of speech events, it becomes interesting to see what happens when more than one human face is present in a frame. In Fig.~\ref{fig:multi_person}, we present the CAMs for selected frames, from the TV show \emph{Andi Mack} and the movie \emph{Dumb and Dumber}, neither seen by the network during training, corresponding to positive class. The frames are particularly selected to have more than one faces present. The figure illustrate that the learned crossmodal embeddings are able to select the active speaker even when more than one face is present on the frame. However, the variance of the heatmaps is not as concentrated as in the case of a single face, shown in Fig.~\ref{fig:single_person}. The sample videos with imposed CAMs can be found in supplementary material.

\textbf{\textit{Importance of stacked LSTMs:}} 
In this work, we enhance the HICA~\cite{icip} architecture with three additional stacked Convolutional-BiLSTM, with an idea that it preserves the spatial and temporal information and achieves hierarchical abstraction. To qualitatively validate the advantages of stacked-ed Convolutional BiLSTM layers, we compare the CAMs at two hierarchical levels, one at the end of 3D convolutional network \emph{(Conv\_CAMS)} and the other at the end of the stacked LSTMs \emph{(LSTM\_CAMs)}. 

\begin{figure}
\centering
\includegraphics[width=0.45\textwidth, keepaspectratio]{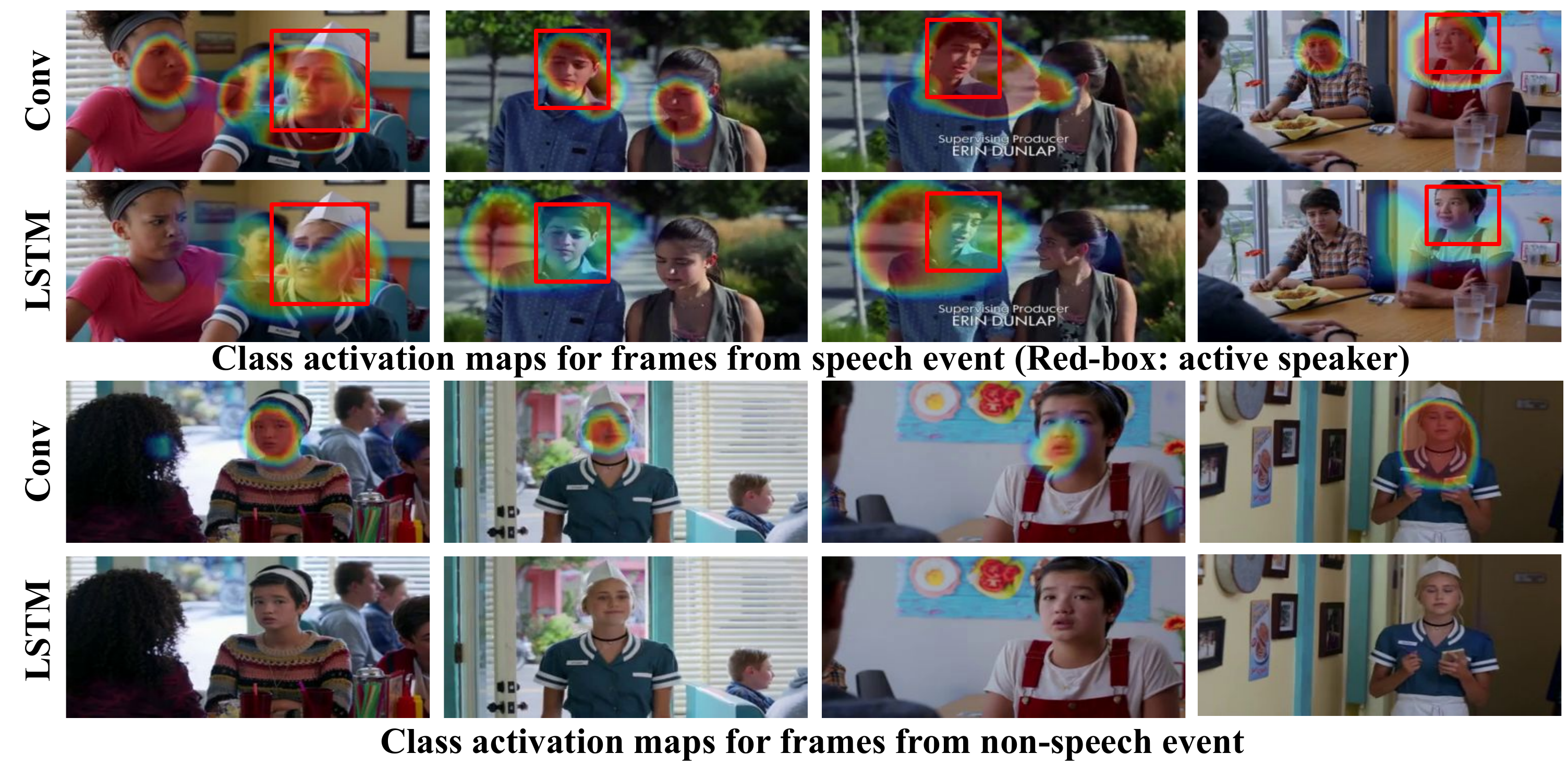}
\caption{Qualitative comparison of CAMs for the last convolutional layer against the last convolutional-LSTM layer for the case of speech and non-speech events.} 
\label{fig:hica_compare}
\end{figure}
In Fig.~\ref{fig:hica_compare}, we present CAMs for the two hierarchical levels under two different scenarios:
\begin{enumerate}
    \item \textbf{Speech event:} We present frames with more than one face present in the frame, from the set of speech events. To make it more informative, we manually marked the active speakers in the frames using a red box. It can be observed that the activations in the case of the \emph{Conv\_CAMs} extend to the non-speaker face as well, while the \emph{LSTM\_CAMs} can correct the activations to concentrate just on the active speaker.
    
    \item \textbf{Non-speech event:} In this scenario, we present the CAMs for the frames corresponding to non-speech events. We observed that the \emph{Conv\_CAMs} are concentrating on the faces visible in the frames irrespective of their activity while the \emph{LSTM\_CAMs} can correct the undesired activations, and are selective to speech events.
\end{enumerate}
It can be inferred that the group of 3D convolutional layers is selecting the available faces in the frames and the stacked LSTMs, as they can observe a longer context, are narrowing down to selecting the active speakers.

\subsection{Quantitative Analysis}
Since we have visually established that the learned cross-modal visual representations can successfully localize the active speakers, now we formally quantify the performance of the embeddings for audio-visual speech event localization. It comprises two tasks, i) voice activity detection in audio modality, and ii) active speaker localization in the visual modality. We further experimentally support the robustness of the visual representations for the localizing task.

\subsubsection{Audio-visual voice activity detection}
\label{subsec:av_vad}
\begin{table}
\centering
\caption{True Positive Rates for FPR=0.315, evaluated for several models for the task of VAD. }
\label{tab:VAD_speech}
\resizebox{0.45\textwidth}{!}{%
\begin{tabular}{cccccc}
\hline
\begin{tabular}[c]{@{}c@{}}Datasets / \\ Models\end{tabular} & resnet90 & tiny 320 & Cldnn & cnn td & ours \\ \hline
\begin{tabular}[c]{@{}c@{}}AVA\\ (Clean Speech)\end{tabular}   & \textbf{0.992}    & 0.965    & 0.985 & 0.983  & 0.92 \\ \hline
\begin{tabular}[c]{@{}c@{}}AVA\\ (Speech + Music)\end{tabular} & 0.787 & 0.632 & 0.906 & 0.917 & \textbf{0.92}  \\ \hline
\begin{tabular}[c]{@{}c@{}}AVA\\ (Speech + Noise)\end{tabular} & 0.944 & 0.826 & 0.922 & 0.93  & \textbf{0.96}  \\ \hline
\begin{tabular}[c]{@{}c@{}}AVA \\ (all)\end{tabular}    & 0.917 & 0.81  & 0.935 & 0.945 & 0.935 \\ \hline
\end{tabular}%
}
\end{table}
We explore the applicability of the cross-modal embeddings for localizing audio speech events. We train a late fusion model for the audio VAD task, in a co-training fashion with the MIL framework, following \S~\ref{subsec:wsod_network}. To validate the performance of the proposed system, we evaluate it on two richly dynamic datasets. The first is the AVA speech dataset \cite{avaSpeech}, which consists of 15 minute video clips from 160 international movies, manually annotated for voice activity. The AVA dataset comprises three different scenarios for speech events, clean speech, speech with music, and speech with noise, which makes it highly challenging. The proposed system is limited to segment level predictions, thus providing binary labels for each 1 sec of the video. To make the AVA dataset compatible with the segment level evaluation, we use majority voting to compute the segment level labels and report performance on validatoin set.
The other dataset we used for evaluation, LLP~\cite{avvp_llp}, consists of $1849$ YouTube videos, each $10 sec$ in duration, and video-level event annotations on the presence or absence of different video events, speech event being one. It offers annotations with granularity being 1 sec, thus are coarse.
% The dataset offers second-wise annotations, as it is designed specifically for weakly-supervised training, and thus are coarse. 
\begin{figure}
\centering
\includegraphics[width=0.35\textwidth,keepaspectratio]{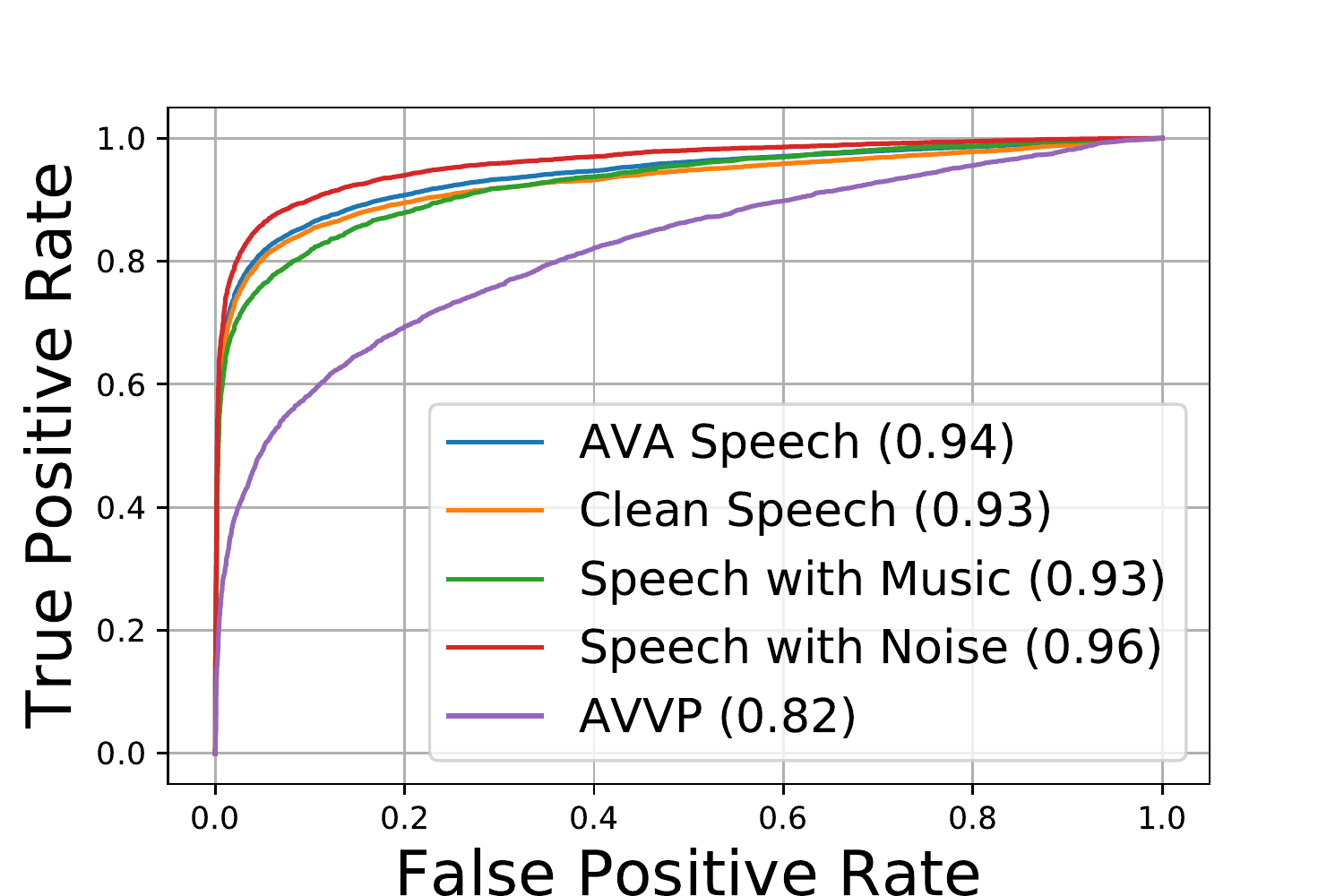}
\caption{Receiver-Operating Characteristics (ROC) for Audio-Visual VAD performance for different datasets.} 
\label{fig:av_vad}
\end{figure}

In Fig.~\ref{fig:av_vad}, we present the receiver-operating characteristics (ROC) curve, along with area under ROC curve, showing the performance of the proposed system across different datasets, evaluated for segment level predictions. In Table~\ref{tab:VAD_speech}, we compare the performance of the system, in terms of true positive rate when false positive rate being 0.315, with the state-of-the-art systems. It can be observed that the proposed audio-visual model outperforms the audio-based state-of-the-art methods in cases when speech is accompanied with music and noise. It can be inferred that the active speaker localizing capability of the cross-modal visual representations, complements the voice activity detection performance when the audio signal is not reliable. For the same experimental setting, we observe a TPR of 0.771 for LLP dataset. Since the dataset is designed for the task of audio-visual event classification, we did not find any reports of evaluations particularly for the speech event to compare against. 

\subsubsection{Active speaker localization}
\begin{figure*}
\centering
\includegraphics[width=\textwidth,keepaspectratio]{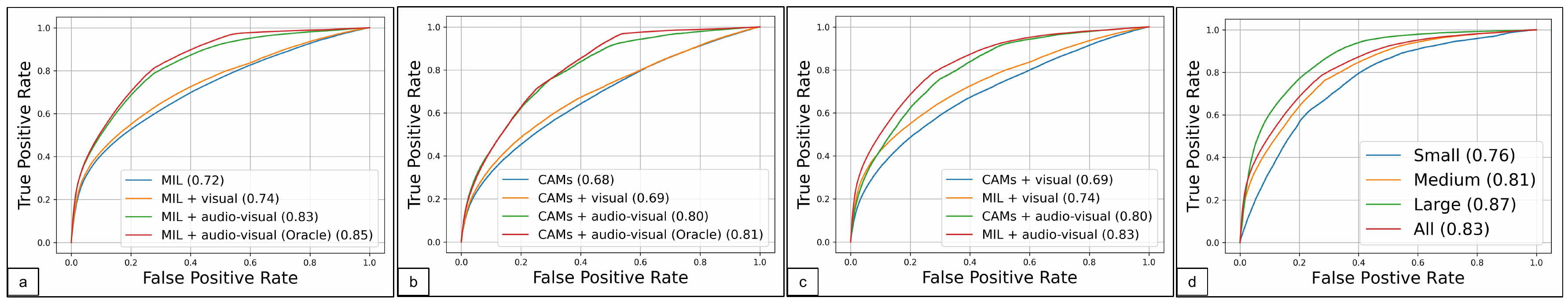}
\caption{Performance (auROC) on AVA dataset a) MIL system posteriors + post processing b) CAMs-assisted system + post processing c) comparison of CAMs and MIL d) with various face sizes.} 
\label{fig:cams_mil_roc}
\end{figure*}
To evaluate the active speaker localization performance of the proposed CAM-assisted system and MIL-based weakly-supervised framework~\S\ref{subsec:wsod_network}, we use the AVA active speaker dataset~\cite{avaActiveSpeaker}. The dataset consists of the same movies as in the AVA speech dataset, described in~\S\ref{subsec:av_vad}, with additional annotations, comprising all the face tracks and manually annotated face-wise active speaker labels (in all $780k$ annotated faces). The movies in the AVA dataset are international movies, shot in an earlier time period, thus are  different from contemporaneous Hollywood movies in terms of cinematography. The authors~\cite{avaActiveSpeaker} also present a fully supervised baseline and reported the performance in terms of area under the ROC curve. The proposed system has never observed any part of the AVA active speaker dataset and we report performance on the validation split (as provided by the authors~\cite{avaActiveSpeaker}), consisting of $33$ movies, to make it comparable.
\begin{table}[tb]
\centering
\caption{Performance (auROC) for active speaker localization on AVA dataset.}
\label{tab:au_ROC}
\resizebox{0.30\textwidth}{!}{%
\begin{tabular}{ccc}
\hline
\begin{tabular}[c]{@{}c@{}}Experimental setup /\\ Methods\end{tabular}             & CAMs & MIL  \\ \hline
\begin{tabular}[c]{@{}c@{}}Visual posteriors\\ (no post processing)\end{tabular}   & 0.68 & 0.72 \\ \hline
\begin{tabular}[c]{@{}c@{}}Visual \\ post processing\end{tabular}                  & 0.69 & 0.74 \\ \hline
\begin{tabular}[c]{@{}c@{}}Visual + audio \\ post processing\end{tabular}          & 0.80 & 0.83 \\ \hline
\begin{tabular}[c]{@{}c@{}}Visual + audio (oracle) \\ post processing\end{tabular} & 0.82 & 0.85 \\ \hline
\begin{tabular}[c]{@{}c@{}}AVA baseline \\ fully supervised system~\cite{avaActiveSpeaker}\end{tabular} & \multicolumn{2}{c}{0.92} \\ \hline
\end{tabular}%
}
\end{table}

In Table~\ref{tab:au_ROC} we report the performance for the CAMs-assisted and the MIL-based system in terms of area under the ROC, as suggested by \cite{avaActiveSpeaker}. For a more elaborate understanding of the performance, we present the ROC curves for all the experiments in Fig~\ref{fig:cams_mil_roc}. We observe the following: 
\begin{itemize}
    \item The CAMs-assisted system, utilizing just the visual signal shows moderate performance for the task of active speaker localization. 
    \item Unlike CAMs, which are purposed for visualizing the network activations, the proposed MIL system observes more informative visual representations for the task of active speaker localization. It shows improved performance for the task compared to the CAMs-assisted method, consistently across all the experiments.
    \item The visual post-processing step, which enforces the uni-speaker constraint for each frame, shows a marginal improvement in the performance, which indicates that the raw posteriors are already highly selective of just one speaker in each frame. 
    \item Further adding audio post-processing step shows a noticeable improvement for both the methods, CAMs-assisted and MIL system. Since the audio post-processing specifically constrains the active speakers in the voice active regions only, the enhanced performance is indicative of the false positives produced by the raw posteriors in the non-speech regions. This can be attributed to the limited information present in the visual modality which solely generates the posteriors. 
    \item We also present the performance of the audio-visual post-processing step, using VAD labels provided by the AVA~\cite{avaActiveSpeaker} annotations. We observe marginal improvement compared to system provided VAD (\S\ref{subsec:wsod_network}), which revalidates the high VAD performance of the proposed system as noted in~\S\ref{subsec:av_vad}.
\end{itemize}
% For a more elaborate view, we present the ROC curves in Fig.~\ref{fig:cams_mil_roc}c, showing that the MIL system boosts the performance over the baseline CAM-based system. The visual-only experiments distinctly demonstrate the improvement provided by the uni-speaker assumption. The VAD post-processing step, for decreasing the false positives, imparts a significant improvement over the visual-only systems and can be easily observed from the ROC curves. The slight degradation in performance when switching from oracle VAD labels to av-VAD system predictions signifies the efficacy of the VAD system. Although the addition of face track information improves performance, it is marginal. 

% \begin{figure}
% \centering
% \includegraphics[width=0.45\textwidth,keepaspectratio]{figures/roc_vs_face_size.pdf}
% \caption{Variation in performance on AVA active speaker dataset against size of faces.} 
% \label{fig:auc_vs_face}
% \end{figure}

% \begin{figure}
% \centering
% \includegraphics[width=0.45\textwidth,keepaspectratio]{figures/roc_with_face_size.pdf}
% \caption{Performance of the MIL setup across different face sizes present in the AVA active speaker dataset.} 
% \label{fig:roc_vs_face}
% \end{figure}

We compare the performance against the baseline setup by \cite{avaActiveSpeaker}. The lower performance against \cite{avaActiveSpeaker} can be primarily attributed to the differences in system setups. Firstly, the network presented in \cite{avaActiveSpeaker} is end-to-end optimized on the AVA dataset in a supervised fashion with audio-visual face-tracks as input and annotated active speaker bounding boxes as labels. While on the other hand, our cross-modal network has been trained in a weakly supervised fashion without encountering annotations for active speakers. Furthermore, it has {\it not} seen the AVA dataset in any form. Secondly, the network output by the cross-modal system is low in spatial resolution ($12\times23$ to be precise), thus features driving the MIL system, using ROI pooling, have limited information for smaller faces. 

To further investigate the effect of the face size on the active speaker localization performance, we present an evaluation based on face box sizes. We divide the set of all the face boxes on the basis of their size into 3 categories: small boxes, occupying less than 2\% of the screen space ($\leq4$ pixels for CAMs ($12\times23$)), medium-sized boxes, occupying 2\%-15\% of the screen space, and large boxes, occupying more than 15\% of the screen space. In Fig.~\ref{fig:cams_mil_roc} we show the ROC curve for the three sets. We observe a clear degradation in performance as the size of the face goes smaller. 
\begin{figure*}
\begin{minipage}{0.75\textwidth}
\centering
\includegraphics[width=0.95\textwidth,keepaspectratio]{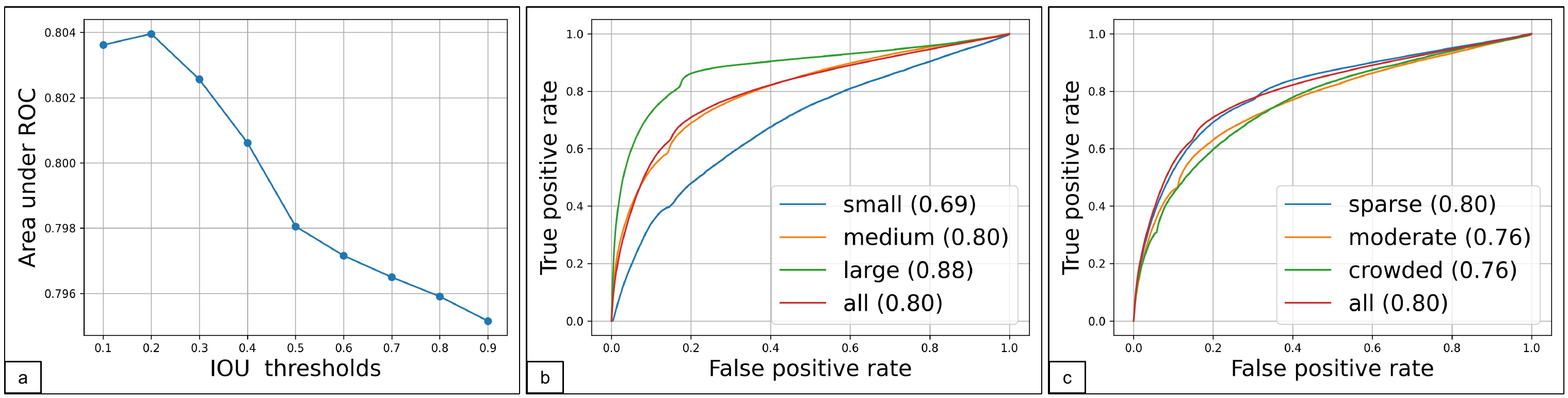}
\caption{Performance (auROC) on AVA dataset with generic object proposals a) with different IOU thresholds. b) for various face sizes c) different number of boxes.} 
\label{fig:od_performance}
\end{minipage}%
\begin{minipage}{0.25\textwidth}
\centering
\includegraphics[width=0.85\textwidth,keepaspectratio]{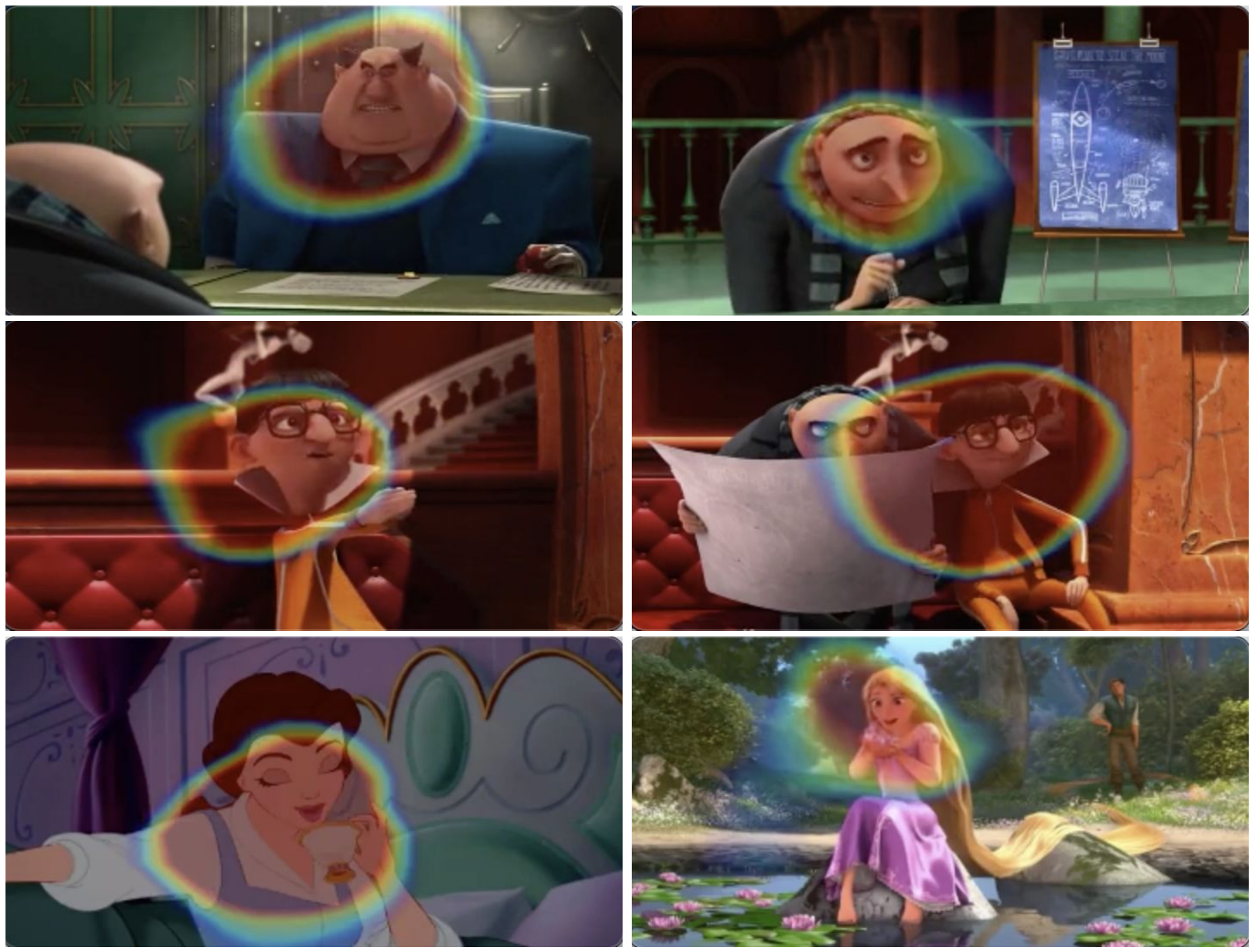}
\caption{Sample of CAMs imposed on frames for animated videos.} 
\label{fig:animated}
\end{minipage}%
\end{figure*}

We stress-test the robustness of the cross-modal MIL setup by generalizing the input proposals to generic object detection boxes, along with the existing face proposals, which we obtain using Faster RCNN~\cite{fastRCNN}. We remove the object boxes which overlap with the face boxes with an $IOU > th$ and label all the new boxes as negative. This increased the number of sample boxes to 4-folds, making it $3M$ for AVA active speaker dataset. The evaluation uses MIL posteriors, followed by the visual and audio post-processing steps. We observe a marginal degradation in performance decreasing from 0.83 to 0.80. This establishes that the system is robust to proposals and can differentiate active human speakers well. 
% In Fig.~\ref{fig:od_performance}a, we present the performance for various values of IOU thresholds. The degradation in the performance is marginal, decreasing from 0.83 to 0.80. This establishes that the system is robust to proposals and can differentiate active human speakers well. 

Greater value of IOU threshold indicates the presence of object boxes with significant overlap with one of the face boxes, but are definitely labeled as negative. Such a scenario leads to more false positives, which is reflected in the form of the decrease in performance with the increase in the IOU threshold, as shown in Fig.~\ref{fig:od_performance}a. For a better understanding, we looked at the variation in performance against the number of boxes present in each frame. For clarity, we divided the dataset into sparse frames with less than 5 boxes, moderate frames with the number of boxes between 5 and 15, and crowded with more than 15 boxes. Although, from Fig.~\ref{fig:od_performance}b, the system performs better with fewer number of boxes in each frame, the difference is marginal, which re-validates the robustness of the system against proposal boxes. We also present the performance against box sizes in Fig.~\ref{fig:od_performance}c, and observed that the performance degrades as the size of the box decreases, consistent with the earlier observation. 

\section{Summary and Future Work}
\label{sec:conclusion}
In this paper, we present a cross-modal framework for learning visual representations, capable of localizing an active speaker in the visual frames. We further formalized a system for active speaker localization, in a weakly supervised manner, requiring no manual annotations. The consistency in the problem formulation for the cross-modal network and the MIL setup makes the system end-to-end trainable. We evaluated the performance of audio-visual speech event localization on the AVA dataset, and demonstrated compelling performance for audio VAD in noisy conditions and showed performance comparable to supervised methods for localizing an active speaker.

The presented system is self-contained in the sense that it can be adapted to any domain in a straightforward manner. To do so, it requires no manual annotations, but just coarse voice activity labels, which can be obtained using the proposed state-of-the-art VAD system. One of the immediate extensions of our work is to adapt the system for animated content for animated character discovery such as illustrated in Fig.~\ref{fig:animated}. %The sneak peek is shown in fig~\ref{fig:animated}.  
Since the system connects the speech to its spatial source in visual frames, it can be extended to jointly model audio and visual modality for diarization. 
% \begin{figure}
% \centering
% \includegraphics[width=0.45\textwidth,keepaspectratio]{figures/animated.pdf}
% \caption{CAMs for selected frames from an animated movie, "Despicable ME"} 
% \label{fig:animated}
% \end{figure}
% \section{Conclusion}
% The conclusion goes here.

% or


% \appendices

% % you can choose not to have a title for an appendix
% % if you want by leaving the argument blank
% \section{}

% % use section* for acknowledgment
% \section*{Acknowledgment}


% \newpage
\ifCLASSOPTIONcaptionsoff
  \newpage
\fi



\bibliographystyle{IEEEtran}
\bibliography{ref}

\begin{thebibliography}{10}
\providecommand{\url}[1]{#1}
\csname url@samestyle\endcsname
\providecommand{\newblock}{\relax}
\providecommand{\bibinfo}[2]{#2}
\providecommand{\BIBentrySTDinterwordspacing}{\spaceskip=0pt\relax}
\providecommand{\BIBentryALTinterwordstretchfactor}{4}
\providecommand{\BIBentryALTinterwordspacing}{\spaceskip=\fontdimen2\font plus
\BIBentryALTinterwordstretchfactor\fontdimen3\font minus
  \fontdimen4\font\relax}
\providecommand{\BIBforeignlanguage}[2]{{%
\expandafter\ifx\csname l@#1\endcsname\relax
\typeout{** WARNING: IEEEtran.bst: No hyphenation pattern has been}%
\typeout{** loaded for the language `#1'. Using the pattern for}%
\typeout{** the default language instead.}%
\else
\language=\csname l@#1\endcsname
\fi
#2}}
\providecommand{\BIBdecl}{\relax}
\BIBdecl

\bibitem{cmi}
K.~Somandepalli, T.~Guha, V.~R. Martinez, N.~Kumar, H.~Adam, and S.~Narayanan,
  ``Computational media intelligence: Human-centered machine analysis of
  media,'' \emph{Proceedings of the IEEE}, pp. 1--20, 2021.

\bibitem{klemen2012current}
J.~Klemen and C.~D. Chambers, ``Current perspectives and methods in studying
  neural mechanisms of multisensory interactions,'' \emph{Neuroscience \&
  Biobehavioral Reviews}, pp. 111--133, 2012.

\bibitem{shams2010crossmodal}
L.~Shams and R.~Kim, ``Crossmodal influences on visual perception,''
  \emph{Physics of life reviews}, pp. 269--284, 2010.

\bibitem{schmiedchen2012crossmodal}
K.~Schmiedchen, C.~Freigang, I.~Nitsche, and R.~R{\"u}bsamen, ``Crossmodal
  interactions and multisensory integration in the perception of audio-visual
  motion—a free-field study,'' \emph{Brain research}, pp. 99--111, 2012.

\bibitem{arandjelovic2017look}
R.~Arandjelovic and A.~Zisserman, ``Look, listen and learn,'' in
  \emph{Proceedings of the IEEE International Conference on Computer Vision},
  2017, pp. 609--617.

\bibitem{objectsthatsound}
------, ``Objects that sound,'' in \emph{Proceedings of the European Conference
  on Computer Vision (ECCV)}, 2018, pp. 435--451.

\bibitem{multisensory_ownes}
A.~Owens and A.~A. Efros, ``Audio-visual scene analysis with self-supervised
  multisensory features,'' in \emph{Proceedings of the European Conference on
  Computer Vision (ECCV)}, 2018, pp. 631--648.

\bibitem{sound_pixel}
H.~Zhao, C.~Gan, A.~Rouditchenko, C.~Vondrick, J.~McDermott, and A.~Torralba,
  ``The sound of pixels,'' in \emph{Proceedings of the European Conference on
  Computer Vision (ECCV)}, 2018, pp. 570--586.

\bibitem{icip}
R.~Sharma, K.~Somandepalli, and S.~Narayanan, ``Toward visual voice activity
  detection for unconstrained videos,'' in \emph{2019 IEEE International
  Conference on Image Processing (ICIP)}.\hskip 1em plus 0.5em minus
  0.4em\relax IEEE, 2019, pp. 2991--2995.

\bibitem{somandepalli2018multimodal}
K.~Somandepalli, V.~Martinez, N.~Kumar, and S.~Narayanan, ``Multimodal
  representation of advertisements using segment-level autoencoders,'' in
  \emph{Proceedings of the 20th ACM International Conference on Multimodal
  Interaction}, 2018, pp. 418--422.

\bibitem{cross-modal(CMRAN)}
\BIBentryALTinterwordspacing
H.~Xu, R.~Zeng, Q.~Wu, M.~Tan, and C.~Gan, ``Cross-modal relation-aware
  networks for audio-visual event localization,'' in \emph{Proceedings of the
  28th ACM International Conference on Multimedia}, ser. MM '20.\hskip 1em plus
  0.5em minus 0.4em\relax New York, NY, USA: Association for Computing
  Machinery, 2020, p. 3893–3901. [Online]. Available:
  \url{https://doi.org/10.1145/3394171.3413581}
\BIBentrySTDinterwordspacing

\bibitem{cross-modal(fake_news)}
\BIBentryALTinterwordspacing
C.~Song, N.~Ning, Y.~Zhang, and B.~Wu, ``A multimodal fake news detection model
  based on crossmodal attention residual and multichannel convolutional neural
  networks,'' \emph{Information Processing and Management}, vol.~58, no.~1, p.
  102437, 2021. [Online]. Available:
  \url{https://www.sciencedirect.com/science/article/pii/S0306457320309304}
\BIBentrySTDinterwordspacing

\bibitem{WSOD_selfthaught}
L.~Bazzani, A.~Bergamo, D.~Anguelov, and L.~Torresani, ``Self-taught object
  localization with deep networks,'' in \emph{2016 IEEE winter conference on
  applications of computer vision (WACV)}.\hskip 1em plus 0.5em minus
  0.4em\relax IEEE, 2016, pp. 1--9.

\bibitem{WSOD_cams}
B.~Zhou, A.~Khosla, A.~Lapedriza, A.~Oliva, and A.~Torralba, ``Learning deep
  features for discriminative localization,'' in \emph{Proceedings of the IEEE
  Conference on Computer Vision and Pattern Recognition (CVPR)}, June 2016.

\bibitem{Grad-CAM}
R.~R. Selvaraju, M.~Cogswell, A.~Das, R.~Vedantam, D.~Parikh, and D.~Batra,
  ``Grad-cam: Visual explanations from deep networks via gradient-based
  localization,'' in \emph{2017 IEEE International Conference on Computer
  Vision (ICCV)}, Oct 2017, pp. 618--626.

\bibitem{WSOD_gradcam++}
A.~Chattopadhay, A.~Sarkar, P.~Howlader, and V.~N. Balasubramanian,
  ``Grad-cam++: Generalized gradient-based visual explanations for deep
  convolutional networks,'' in \emph{2018 IEEE Winter Conference on
  Applications of Computer Vision (WACV)}, 2018, pp. 839--847.

\bibitem{WSOD_wsddn}
H.~Bilen and A.~Vedaldi, ``Weakly supervised deep detection networks,'' in
  \emph{2016 IEEE Conference on Computer Vision and Pattern Recognition
  (CVPR)}, 2016, pp. 2846--2854.

\bibitem{WSOD_midn}
P.~Tang, X.~Wang, X.~Bai, and W.~Liu, ``Multiple instance detection network
  with online instance classifier refinement,'' in \emph{Proceedings of the
  IEEE Conference on Computer Vision and Pattern Recognition}, 2017, pp.
  2843--2851.

\bibitem{WSOD_cmidn}
Y.~Gao, B.~Liu, N.~Guo, X.~Ye, F.~Wan, H.~You, and D.~Fan, ``C-midn: Coupled
  multiple instance detection network with segmentation guidance for weakly
  supervised object detection,'' in \emph{Proceedings of the IEEE/CVF
  International Conference on Computer Vision}, 2019, pp. 9834--9843.

\bibitem{WSOD_cmil}
F.~Wan, C.~Liu, W.~Ke, X.~Ji, J.~Jiao, and Q.~Ye, ``C-mil: Continuation
  multiple instance learning for weakly supervised object detection,'' in
  \emph{Proceedings of the IEEE/CVF Conference on Computer Vision and Pattern
  Recognition}, 2019, pp. 2199--2208.

\bibitem{everingham2006hello}
M.~Everingham, J.~Sivic, and A.~Zisserman, ``Hello! my name is...
  buffy''--automatic naming of characters in tv video.'' in \emph{BMVC},
  vol.~2, no.~4, 2006, p.~6.

\bibitem{outoftime}
J.~S. Chung and A.~Zisserman, ``Out of time: automated lip sync in the wild,''
  in \emph{Asian conference on computer vision}.\hskip 1em plus 0.5em minus
  0.4em\relax Springer, 2016, pp. 251--263.

\bibitem{JayChakroborty}
P.~Chakravarty, S.~Mirzaei, T.~Tuytelaars, and H.~Van~hamme, ``Who's speaking?
  audio-supervised classification of active speakers in video,'' in
  \emph{Proceedings of the 2015 ACM on International Conference on Multimodal
  Interaction}, 2015, pp. 87--90.

\bibitem{jayactive}
P.~Chakravarty, J.~Zegers, T.~Tuytelaars, and H.~Van~hamme, ``Active speaker
  detection with audio-visual co-training,'' in \emph{Proceedings of the 18th
  ACM International Conference on Multimodal Interaction}, 2016, pp. 312--316.

\bibitem{avaActiveSpeaker}
J.~Roth, S.~Chaudhuri, O.~Klejch, R.~Marvin, A.~Gallagher, L.~Kaver,
  S.~Ramaswamy, A.~Stopczynski, C.~Schmid, Z.~Xi \emph{et~al.},
  ``Ava-activespeaker: An audio-visual dataset for active speaker detection,''
  \emph{arXiv preprint arXiv:1901.01342}, 2019.

\bibitem{ava_multitask}
Y.-H. Zhang, J.~Xiao, S.~Yang, and S.~Shan, ``Multi-task learning for
  audio-visual active speaker detection.''

\bibitem{ava_naver}
J.~S. Chung, ``Naver at activitynet challenge 2019--task b active speaker
  detection (ava),'' \emph{arXiv preprint arXiv:1906.10555}, 2019.

\bibitem{ava_asdcontext}
J.~L. Alcazar, F.~Caba, L.~Mai, F.~Perazzi, J.-Y. Lee, P.~Arbelaez, and
  B.~Ghanem, ``Active speakers in context,'' in \emph{Proceedings of the
  IEEE/CVF Conference on Computer Vision and Pattern Recognition (CVPR)}, June
  2020.

\bibitem{asd_selfsupervised}
T.~Afouras, A.~Owens, J.~S. Chung, and A.~Zisserman, ``Self-supervised learning
  of audio-visual objects from video,'' \emph{arXiv preprint arXiv:2008.04237},
  2020.

\bibitem{ssl_old1}
\BIBentryALTinterwordspacing
J.~Hershey and J.~Movellan, ``Audio vision: Using audio-visual synchrony to
  locate sounds,'' in \emph{Advances in Neural Information Processing Systems},
  S.~Solla, T.~Leen, and K.~M\"{u}ller, Eds., vol.~12.\hskip 1em plus 0.5em
  minus 0.4em\relax MIT Press, 2000. [Online]. Available:
  \url{https://proceedings.neurips.cc/paper/1999/file/b618c3210e934362ac261db280128c22-Paper.pdf}
\BIBentrySTDinterwordspacing

\bibitem{ssl_harmony}
Z.~Barzelay and Y.~Y. Schechner, ``Harmony in motion,'' in \emph{2007 IEEE
  Conference on Computer Vision and Pattern Recognition}, 2007, pp. 1--8.

\bibitem{ssl_old2}
J.~W. Fisher~III, T.~Darrell, W.~T. Freeman, and P.~A. Viola, ``Learning joint
  statistical models for audio-visual fusion and segregation,'' in
  \emph{Advances in neural information processing systems}, 2001, pp. 772--778.

\bibitem{ssl_soundmotion}
H.~Zhao, C.~Gan, W.-C. Ma, and A.~Torralba, ``The sound of motions,'' in
  \emph{Proceedings of the IEEE/CVF International Conference on Computer
  Vision}, 2019, pp. 1735--1744.

\bibitem{ssl_od}
T.~Afouras, Y.~M. Asano, F.~Fagan, A.~Vedaldi, and F.~Metze, ``Self-supervised
  object detection from audio-visual correspondence,'' \emph{arXiv preprint
  arXiv:2104.06401}, 2021.

\bibitem{training_hard}
W.~Wang, D.~Tran, and M.~Feiszli, ``What makes training multi-modal
  classification networks hard?'' in \emph{Proceedings of the IEEE/CVF
  Conference on Computer Vision and Pattern Recognition}, 2020, pp.
  12\,695--12\,705.

\bibitem{hermans2013training}
M.~Hermans and B.~Schrauwen, ``Training and analysing deep recurrent neural
  networks,'' in \emph{Advances in neural information processing systems},
  2013, pp. 190--198.

\bibitem{WSOD_instance}
Z.~Ren, Z.~Yu, X.~Yang, M.-Y. Liu, Y.~J. Lee, A.~G. Schwing, and J.~Kautz,
  ``Instance-aware, context-focused, and memory-efficient weakly supervised
  object detection,'' in \emph{Proceedings of the IEEE/CVF conference on
  computer vision and pattern recognition}, 2020, pp. 10\,598--10\,607.

\bibitem{rajat_vad}
R.~Hebbar, K.~Somandepalli, and S.~Narayanan, ``Robust speech activity
  detection in movie audio: Data resources and experimental evaluation,'' in
  \emph{ICASSP 2019-2019 IEEE International Conference on Acoustics, Speech and
  Signal Processing (ICASSP)}.\hskip 1em plus 0.5em minus 0.4em\relax IEEE,
  2019, pp. 4105--4109.

\bibitem{fastRCNN}
X.~Wang, A.~Shrivastava, and A.~Gupta, ``A-fast-rcnn: Hard positive generation
  via adversary for object detection,'' in \emph{Proceedings of the IEEE
  conference on computer vision and pattern recognition}, 2017, pp. 2606--2615.

\bibitem{linear_softmax}
Y.~Wang, J.~Li, and F.~Metze, ``A comparison of five multiple instance learning
  pooling functions for sound event detection with weak labeling,'' in
  \emph{ICASSP 2019-2019 IEEE International Conference on Acoustics, Speech and
  Signal Processing (ICASSP)}.\hskip 1em plus 0.5em minus 0.4em\relax IEEE,
  2019, pp. 31--35.

\bibitem{audio-set}
J.~F. Gemmeke, D.~P.~W. Ellis, D.~Freedman, A.~Jansen, W.~Lawrence, R.~C.
  Moore, M.~Plakal, and M.~Ritter, ``Audio set: An ontology and human-labeled
  dataset for audio events,'' in \emph{Proc. IEEE ICASSP 2017}, New Orleans,
  LA, 2017.

\bibitem{avaSpeech}
S.~Chaudhuri, J.~Roth, D.~P. Ellis, A.~Gallagher, L.~Kaver, R.~Marvin,
  C.~Pantofaru, N.~Reale, L.~G. Reid, K.~Wilson \emph{et~al.}, ``Ava-speech: A
  densely labeled dataset of speech activity in movies,'' \emph{arXiv preprint
  arXiv:1808.00606}, 2018.

\bibitem{avvp_llp}
Y.~Tian, D.~Li, and C.~Xu, ``Unified multisensory perception: weakly-supervised
  audio-visual video parsing,'' \emph{arXiv preprint arXiv:2007.10558}, 2020.

\end{thebibliography}
\end{document}